\documentclass[final]{cvpr}

\usepackage{times}
\usepackage{epsfig}
\usepackage{graphicx}
\usepackage{amsmath}
\usepackage{amssymb}
\usepackage{multirow}

\usepackage[pagebackref=true,breaklinks=true,colorlinks,bookmarks=false]{hyperref}

\author{Gangming Zhao,
        Baolian Qi,
        Jinpeng Li,
        \thanks{\textit{Gangming Zhao and Baolian Qi contributed equally to this work.}}
\thanks{Gangming Zhao is with the Department of Computer Science, The University of Hong Kong, Hong Kong.}
\thanks{Qibao Lian and Jinpeng Li are with University of Chinese Academy of Sciences, Beijing, China}
}

\begin{document}

\title{Cross Chest Graph for Disease Diagnosis with Structural \\ Relational Reasoning} 

\maketitle
\thispagestyle{empty}
\begin{abstract}
Locating lesions is important in the computer-aided diagnosis of X-ray images. However, box-level annotation is time-consuming and laborious. How to locate lesions accurately with few, or even without careful annotations is an urgent problem. Although several works have approached this problem with weakly-supervised methods, the performance needs to be improved. One obstacle is that general weakly-supervised methods have failed to consider the characteristics of X-ray images, such as the highly-structural attribute. We therefore propose the Cross-chest Graph (CCG), which improves the performance of automatic lesion detection by imitating doctor's training and decision-making process. CCG models the intra-image relationship between different anatomical areas by leveraging the structural information to simulate the doctor's habit of observing different areas. Meanwhile, the relationship between any pair of images is modeled by a knowledge-reasoning module to simulate the doctor's habit of comparing multiple images. We integrate intra-image and inter-image information into a unified end-to-end framework. Experimental results on the NIH Chest-14 database (112,120 frontal-view X-ray images with 14 diseases) demonstrate that the proposed method achieves state-of-the-art performance in weakly-supervised localization of lesions by absorbing professional knowledge in the medical field.
\end{abstract}

\section{Introduction}
Chest radiographs are a type of medical images that can be conveniently acquired for disease diagnosis. With the rapid development of deep learning, automatic disease detection in chest X-ray images has become an important task in the computer-aided diagnosis.
\begin{figure}[t]
\begin{center}
\includegraphics[width=1\linewidth]{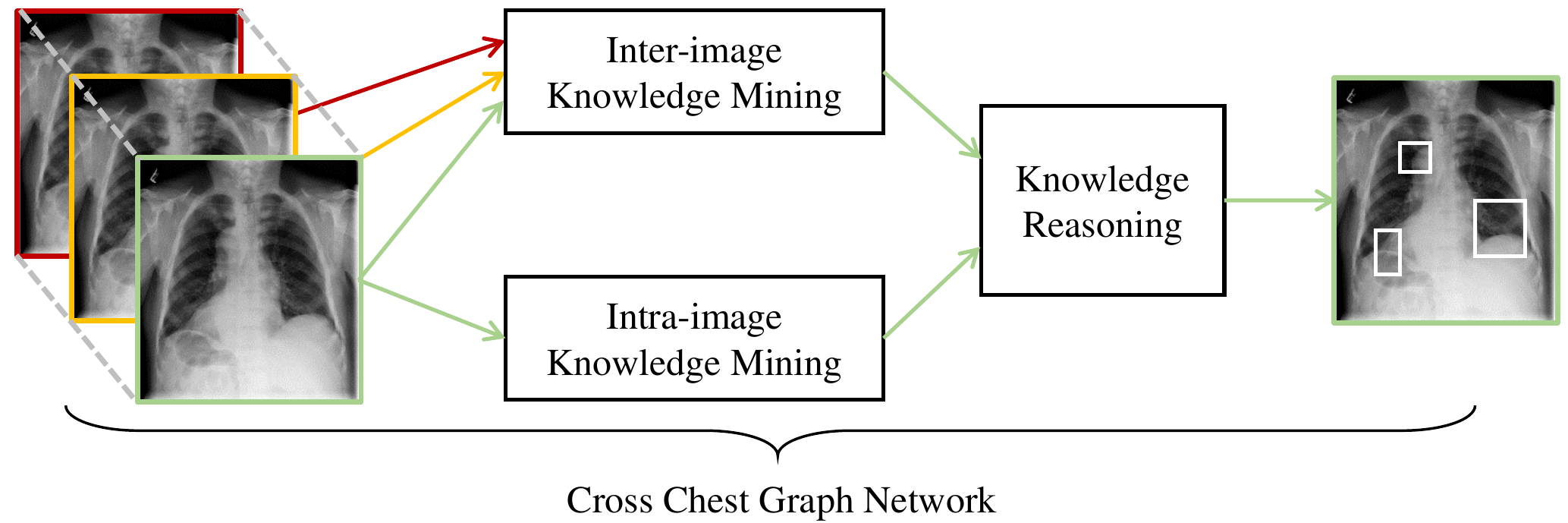}
\end{center}
\caption{CCG network models the intra-image relationship between different anatomical areas by leveraging the structural information to simulate the doctor's habit of observing different areas. Meanwhile, the relationship between any pair of images is modeled by a knowledge-reasoning module to simulate the doctor's habit of comparing multiple images.}\label{fig:intro}
\end{figure}
Deep convolutional neural networks (DCNN) have been widely applied in many computer vision tasks, such as image classification~\cite{he2016deep,russakovsky2015imagenet} , object detection~\cite{girshick2015fast,SPPnet,NIPS2015_5638,redmon2016you,liu2016ssd} and semantic segmentation~\cite{long2015fully,ronneberger2015u}. To achieve good performance in these tasks, substantial images with careful annotations are needed. Encouraged by the success of DCNN in computer vision, some researches have directly applied DCNN models to analyze the medical images but cannot achieve the same performance as in the natural images. The reasons lie in two folds: 1. it is expensive to acquire accurate localization or classification labels in chest X-ray images. 2. there exists much professional knowledge in medical images that DCNN cannot exploit well. Therefore, how to exploit the professional knowledge into DCNN models for solving these two questions still opens a fully challenging problem. Our work transfers the knowledge into DCNN models to reduce the problem of shortage of carefully annotated images.

Recent work paid much attention to utilize professional knowledge of chest X-ray images into DCNN frameworks. However, they just proposed a simple fused strategy to embed low-level information of chest X-ray into models, such as Liu et al.~\cite{liu2019align} utilized contrastive learning to provide more localization information with the help of healthy images. Zhao et al.~\cite{zhao2020contralaterally} proposed to exploit the contralateral information of chest X-ray via a simple fusion module. These methods only exploit the apparent information of chest-Xray images.  They all overlooked the inner structure information of chest X-rays. Therefore, they cannot apply their methods into real applications.

In this paper, we propose a Cross Chest Graph Network (CCG-Net) as shown in Fig \ref{fig:intro}, which firstly utilizes deep expert knowledge to automatical detect disease in chest X-ray images. We have known that medical experts have much experience in finding out disease and how to treat patients. In fact, the actions of medical experts consist of two phases: training and decision-making processes. They pay much time to learn distinguish disease and embed their experience into the decision process. During the training process, experts would like to observe different areas and find out the relationship between any pair of images. Our CCG-Net aims to model the observation way by a knowledge-reasoning module to simulate the doctor's habit of comparing multiple images. Then we integrate intra-image and inter-image information into a unified end-to-end framework.

Inspired from the experience of medical experts, our proposed CCG-Net consists of four modules, 1. an end-to-end framework for deciding where and what is a disease, 2. a inter-image relation module, which formulates the training process of medical experts, to compare multiple images, 3. a intra-image knowledge learning module, which builds the local relation graph for different patches of chest X-ray images. Due to their highly structured property, every chest X-ray image can be divided into several patches, we build a patch-wise relation graph on them, 4. a knowledge reasoning module, which excavates the inner knowledge from cross-image structural features. The last three operations (2, 3, and 4) are similar to medical experts' training process, which learn intra-image and inter-image information to gain professional knowledge. The first operation embeds the learned knowledge into DCNN frameworks leading to better disease diagnosis models.
Above all, our contribution consists of three folds:
\begin{itemize}
  \item We propose CCG-Net, which is the first to formulate the medical experts' training process by building relation graphs in the intra-image and inter-image information of chest X-ray images. More generally, it provides inspiration to address medical vision tasks with much professional knowledge like in chest X-ray images.
  \item We divide the experts' professional actions into two stages including training and decision-making processes. In addition, we utilize intra-image and inter-image relation to learn much professional knowledge that would be embedded in an end-to-end detection framework.
  \item We achieve state-of-the-art results on the localization of NIH ChestX-ray14.
\end{itemize}

\section{Related Work}

\subsection{Disease Detection}
Object detection is one of the most important computer vision tasks, aiming to localize and classify. Due to their strong feature representation ability, DCNN achieved much progress in object detection tasks. For detection tasks, DCNN methods consist of two style framework: 1. two-stage models, such as RCNN series~\cite{NIPS2015_5638}, 2. one-stage models, such as YOLO~\cite{redmon2016you} and SSD~\cite{liu2016ssd}. However, for disease detection, because of the shortage in careful annotations, traditional detection framework cannot directly be applied in chest X-ray images. Besides, since there is much distortion caused by other chest X-ray tissues, such low contrast also causes the difficulty of disease finding.

Weakly supervised object detection (WSOD) can be considered as an effective method to solve these problems. Based on CAM~\cite{zhou2016learning}, researchers proposed many techniques to use only image-level labels to detect objects. Although there is no enough detection supervision, WSOD still achieved much progress. However, researchers still face a big challenge when it comes to disease detection in medical images. the existence of much professional knowledge greatly limits the development of the applications of DCNN in medical fields. Therefore, in this paper, we are inspired by the experts' learning and decision processes to propose CCG-Net, which not only exploits a larger amount of knowledge in chest X-ray images but also builds a unified framework to detect disease in an end-to-end style.

\subsection{Knowledge-based Disease Diagnosis}
Automatical disease diagnosis is a key problem in medical fields. However, due to the shortage of careful annotations and the existence of much professional knowledge, DCNN methods cannot achieve a good performance in medical tasks, especially such a tough problem: disease detection in chest X-ray images. To exploit medical knowledge and embed it into DCNN frameworks, researchers paid much effort to utilize medical experts' experience for disease diagnosis. Wang et al.~\cite{wang2017chestx} firstly proposed a carefully annotated chest X-ray dataset and led to a series of work that focuses on using image-level labels to localize the disease. Li et al.~\cite{li2018thoracic} integrated classification and localization in a whole framework with two multi instance-level losses and performed better. Liu et al.~\cite{liu2019align} improved their work to propose contrastive learning of paired samples, which utilizes healthy images to provide more localization information for disease detection. Zhao et al.~\cite{zhao2020contralaterally} proposed to utilize the symmetry information in a chest X-ray to improve the disease localization performance.
\begin{figure*}[t]
\begin{center}
\includegraphics[width=0.85\linewidth]{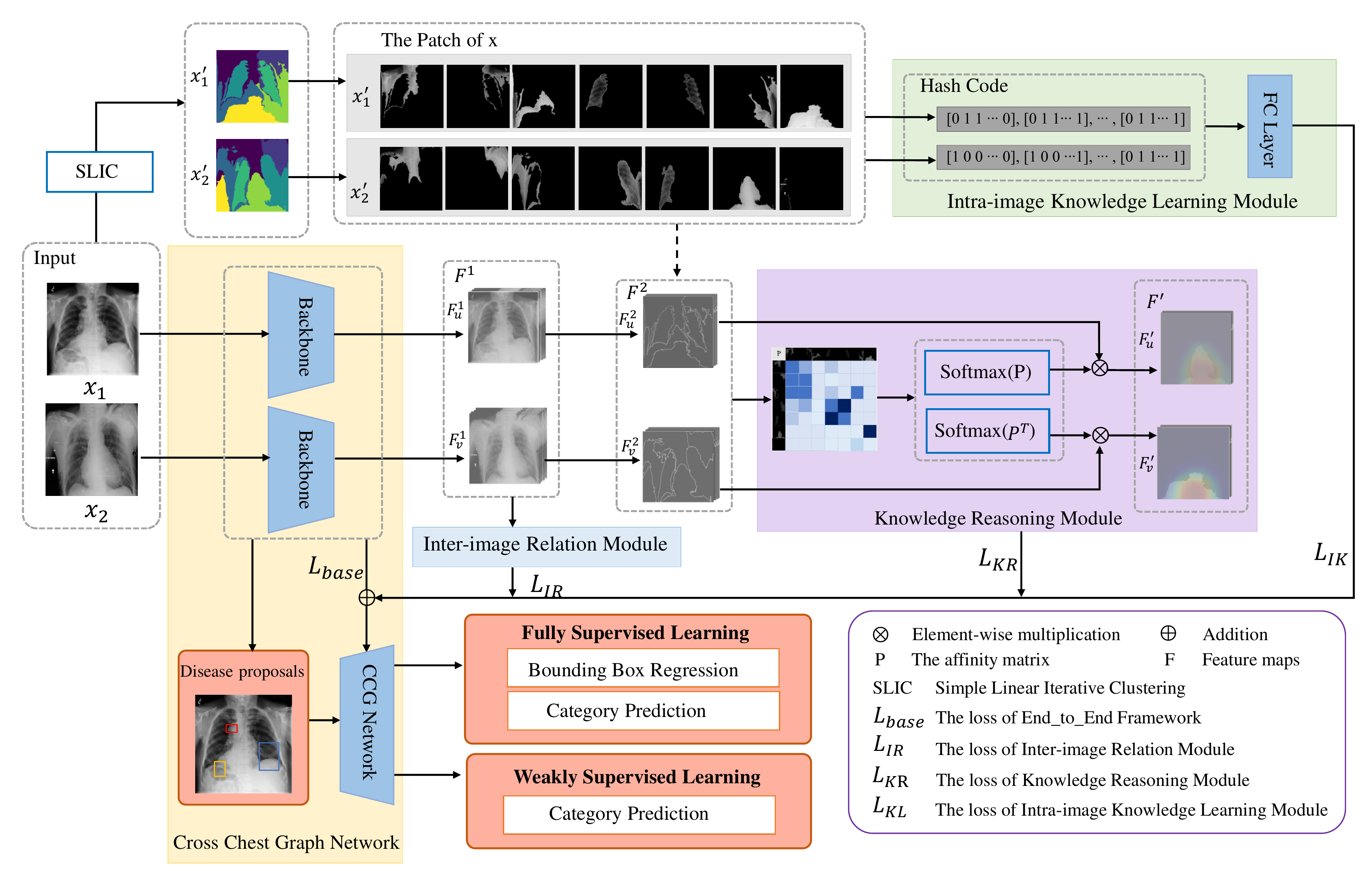}
\end{center}
\caption{The network consists of four modules: 1. an end-to-end framework for disease detection under weakly-supervised settings, 2. the inter-image relation module among different samples, 3. the intra-image knowledge learning module based on the thoracic spatial structure, 4. the knowledge reasoning module mining cross-image structural features. Our four modules are tightly related and can be easily integrated into an end-to-end framework.}\label{fig:framework}
\end{figure*}
Besides, many works applied relation knowledge models to chest X-ray diagnosis. Ypsilantis et al.~\cite{ypsilantis2017learning}, Pesce et al.~\cite{pesce2017learning}, and Guan et al.~\cite{guan2018diagnose} proposed to build a relation attention model fusing DCNN models achieved much progress. Li et al.~\cite{li2019knowledge} proposed a knowledge-graph based on medical reports and images to determine the dependencies among chest X-ray images. Cheng et al.~\cite{liu2020rethinking} also proposed a new total strongly supervised dataset for tuberculosis detection. However, they all overlooked the structural relation among chest X-ray images. In this paper, we propose to build a structural, relational graph for disease detection under weakly supervised scenarios in chest X-ray images. Specifically, we build the global and local graph in chest X-ray via three modules: 1. a inter-image relation module, 2. a intra-image knowledge learning module, 3. knowledge reasoning module. Furthermore, we integrate three modules into an end-to-end framework to jointly train our network. Our proposed three relational modules provide better supervision since we exploit the local structural knowledge and global relation among different samples.

\section{Method}
\subsection{Overview}

Given the images $X = \{x_{1}, x_{2},...,x_{n}\}$. Our proposed framework consists of four modules:
\begin{itemize}
  \item The End to End Framework is to localize and classify the disease in chest X-ray images. In our paper, we utilize the same multi-instance level losses used in~\cite{liu2019align} and~\cite{li2018thoracic}.
  \item Inter-image Relation Module, which includes a learnable matrix $G \in R^{n \times n}$. We also use a contrast-constrained loss to share similar information of $X$ and exploit their contrasted structural knowledge. We build a cross-sample graph for them to exploit the dependencies among different samples. The graph $G \in R^{n\times n}$ is to build the inter-image relation among sampled samples, which is a learnable matrix, and every element is initialized by $\frac{1}{n}$. The element $g_{ij}$ of $G$, $i,j \in \{1,2,...,n\}$, represents the similarity wight of images $x_{i}$ and $x_{j}$.
  \item Intra-image Knowledge Learning, which firstly acquires patch-wise features of different images. Then the network can achieve a new image graph via building a structural knowledge-based module. We denote this graph as $G_{k} \in R^{n \times n}$. Assumed that the number of patches are $|p_{i}|$ and $|p_{j}|$ of images $x_{i}$ and $x_{j}$. The graph $G_{k}$ would be calculated on using the graph $G_{l} \in R^{|p_{i}| \times |p_{j}|}$, which learns the relationship between different paired patches of images.
  \item Knowledge Reasoning Module, which is based on cross image structural knowledge. When we get the whole structural information of different images, we will utilize it to reason the inner structural dependencies among different patches in different images.
\end{itemize}

\subsection{End to End Framework}
The end to end framework is to localize and classify the disease in chest X-ray images in a coarse-grained style. More specifically, the input images $X = \{x_{1}, x_{2},...,x_{n}\}$ of the module are resized to $512 \times 512$.
ResNet-50 pre-trained from the ImageNet dataset is adopted as the backbone for this module. We use the feature map $F$ after C5 (last convolutional output of 5th-stage), which is 32 times down-sampled from the input image, and of size $2048\times 16\times 16$. Each grid in the feature map denotes the existent probability of disease. We pass $F$ through two $1\times 1$ convolutional layers and a sigmoid layer to obtain the class-aware feature map $\emph{P}$ of size $C\times H \times W$, where $C$ is the number of classes. Then we follow the paradigm used in~\cite{liu2019align}, computing losses and making predictions in each channel for the corresponding class.
For images with box-level annotations, if the grid in the feature map overlaps with the projected ground truth box, we assign
label 1 to the grid. Otherwise, we assign 0 to it. Therefore, we use the binary cross-entropy loss as used in ~\cite{liu2019align} for each grid:
\begin{equation}
L^k_i(\emph{P}) = \sum_j -y_{ij}^k \log(p_{ij}^k) - \sum_j (1-y_{ij}^k) \log(1-p_{ij}^k)
\end{equation}
where $k$, $i$, and $j$ are the index of classes, samples, and grids respectively. $y^k_{ij}$ denotes the target label of the grid and
$p^k_{ij}$ denotes the predicted probability of the grid.

For images with only image-level annotations, we use the MIL loss used in~\cite{li2018thoracic}.
\begin{equation}
\begin{split}
L^k_i(\emph{P}) = -&y^k_i\log(1-\prod_{j} (1-p^k_{ij}))\\
                -&(1-y^k_i)\log(\prod_{j} (1-p^k_{ij}))
\end{split}
\end{equation}
where $y^k_i$ denotes the target label of the image. For this end to end framework, the whole loss $L_{base}$ as shown in Fig. \ref{fig:framework}, is formulated as follows.
\begin{equation}
\begin{split}
L_{base} = \sum_i \sum_k \lambda^k_i \beta_B L^k_i(\emph{P}) + (1-\lambda^k_i) L^k_i(\emph{P})
\end{split}
\end{equation}
where $\lambda^k_i \in {0,1}$ denotes if the $k_{th}$ class in the $i_{th}$ sample has box annotation,
and $\beta_B$ is the balance weight of the two losses and is set to 4.
\subsection{Inter-image Relation Module}
Inter-image relation is formulated as a learnable matrix $G \in R^{n \times n}$. A contrast-constrained loss is used to share similar information of $X$ and exploit their contrasted structural knowledge, as following equation.
\begin{equation}
\begin{split}
L_{IR} = \frac{\sum_{(u, v) \in G}G(u, v)D(F_{u}, F_{v})}{n \times n}
\end{split}
\end{equation}
$D(\cdot)$ is the distance metric function, where it is a Euclidean distance. $F_{u}$ and $F_{v}$ means the feature map after C5 of the image $x_{u}$ and $x_{v}$.
We build a cross-sample graph for them to exploit the dependencies among different samples. The graph $G \in R^{n\times n}$ is to build the inter-image relation among sampled samples, which is a learnable matrix, and every element is initialized by $\frac{1}{n}$. The element $g_{ij}$ of $G$, $i,j \in \{1,2,...,n\}$, represents the similarity wight of images $x_{i}$ and $x_{j}$. G is adaptively adjusted during training processes and changes with diverse inputs to exploit the relationship fully.
\subsection{Intra-image Knowledge Learning}
Intra-image Knowledge Learning, which firstly utilizes Simple linear iterative clustering (SLIC)~\cite{achanta2010slic}, a super-pixel method to generate the patches for different images. Assumed that the patches of the image $x_{i}$ is $p_{i} = {p^{i}_{1}, p^{i}_{2},...,p^{i}_{m}}$. Then the network can achieve a new image graph via building a structural knowledge-based module with the help of $p_{i}, i \in {1,2,...,n}$. We denote the graph as $G_{k} \in R^{n \times n}$, which is the intra-image graph between paired images $x_{i}$ and $x_{j}$. The graph $G_{k}$ is calculated on using the graph $G_{l} \in R^{|p_{i}| \times |p_{j}|}$, which learns the dependencies among different paired patches of images. Then the same contrast-constrained loss using this graph to provide more structural knowledge for the whole framework.
\begin{equation}
\begin{split}
L_{IK} = \frac{\sum_{(u, v) \in G_{k}}G_{k}(u, v)D(F_{u}, F_{v})}{n \times n}
\end{split}
\end{equation}
\begin{equation}
\begin{split}
G_{k} = W_{l}(G_{l})
\end{split}
\end{equation}
Where, $W_{l}$ is a fully connected layer and
\begin{equation}
\begin{split}
 G_{l}(l, p) = D^{'}(H_{l}, H^{'}_{p}), l \in {1,2,...,|p_{i}|}, p \in {1,2,...,|p_{j}|}
\end{split}
\end{equation}
$H_{l}$ is the hash code~\cite{2000Robust} of the patch $p^{i}_{l}$ in the image $x_{i}$ and $H^{'}_{p}$ is the hash code of the patch $p^{j}_{p}$ in the image $x_{j}$. $D^{'}(\cdot)$ is the Hamming distance.

\subsection{Knowledge Reasoning Module}

In addition to previous efforts to focus on information in a whole image, we also explored the value of cross-image semantic relations in the medical object. The correlations between patches across images are emphasized, especially, the correlations between corresponding patches in two images.

Knowledge Reasoning Module focuses on the correlations of two images. After getting the feature map $F_u$ and $F_v$ of the images, the affinity matrix $P$ is firstly calculated between $F_{u}$ and $F_{v}$.
\begin{equation*}
P = F^\mathrm{T}_uW_PF_v \in \mathbb{R}^{HW\times HW}
\end{equation*}
where the feature map $ F_u \in \mathbb{R}^{C \times HW}$ and $ F_v \in \mathbb{R}^{C \times HW}$ are flattened into matrix formats, and $ W_P \in \mathbb{R}^{C \times C}$ is a learnable matrix. The affinity matrix $ P $ represents the similarity of all pairs of patches in $F_u$ and $F_v$.

Then $ P $ is normalized column-wise to get the attention map of $F_u$ for each patch in $F_v$ and row-wise to get the attention map of $F_v$ for each patch in $F_u$.

\begin{equation*}
F^{'}_u = F_u softmax(P) \in \mathbb{R}^{C \times HW}
\end{equation*}
\begin{equation*}
F^{'}_v = F_v softmax(P^\mathrm{T}) \in \mathbb{R}^{C \times HW}
\end{equation*}
where $softmax(P)$ and $softmax(P^\mathrm{T})$ pay attention to the similar patches of the feature map $F_u$ and $F_v$ respectively. Therefore, they can be used to enhance $F_u$ and $F_v$ respectively, so that similar patches in $F_u$ and $F_v$ are highlighted.

The cross-image method can extract more contextual information between images than using a single image. This module exploits the context of other related images to improve the reasoning ability of the feature map, which is beneficial to the localization and classification of disease in chest X-ray images. Furthermore, we exploit the enhanced feature map to calculate the new similarity between the paired images to gain a more strong supervisor.
\begin{equation}
\begin{split}
L_{KR} = \frac{\sum_{(u, v) \in G_{k}^{'}}G_{k}^{'}(u, v)D(F^{'}_{u}, F^{'}_{v})}{n \times n}
\end{split}
\end{equation}
\begin{table*}
\small
\begin{center}
\resizebox{0.8\textwidth}{!}{
\begin{tabular}{|l|c c c c c c c c c c|}
\hline
T (IoU) & Models & Atelectasis& Cardiomegaly & Effusion & Infiltration & Mass & Nodule & Pneumonia & Pneumothorax & Mean \\
\hline\hline
 	\multirow{3}{*}{0.3} & X, Wang~\cite{wang2017chestx} & 0.24 & 0.46 & 0.30 & 0.28 & 0.15 & 0.04 & 0.17 & 0.13 & 0.22 \\
	& Z, Li~\cite{li2018thoracic} & 0.36 & \textbf{0.94} & 0.56 & 0.66 & 0.45 & 0.17 & 0.39 & \textbf{0.44} & 0.49 \\
	& J, Liu~\cite{liu2019align} & \textbf{0.53} & 0.88 & 0.57 & 0.73 & \textbf{0.48} & 0.10 & 0.49 & 0.40 & 0.53 \\
	& Ours & 0.44 & 0.86 & \textbf{0.68} & \textbf{0.84} & 0.47 & \textbf{0.29} & \textbf{0.67} & 0.40 & \textbf{0.60} \\

\hline
	\multirow{3}{*}{0.5}& X, Wang~\cite{wang2017chestx} & 0.05 & 0.18 & 0.11 & 0.07 & 0.01 & 0.01 & 0.03 & 0.03 & 0.06 \\
	& Z, Li~\cite{li2018thoracic} & 0.14 & 0.84 & 0.22 & 0.30 & 0.22 & 0.07 & 0.17 & 0.19  & 0.27 \\
	& J, Liu~\cite{liu2019align} & \textbf{0.32} & 0.78 & 0.40 & 0.61 & 0.33 & 0.05 & 0.37 & 0.23 & 0.39 \\
	& Ours & 0.27 & \textbf{0.86} & \textbf{0.48} & \textbf{0.72} & \textbf{0.53} & \textbf{0.14} & \textbf{0.58} & \textbf{0.35} & \textbf{0.49} \\
\hline
 	\multirow{3}{*}{0.7} & X, Wang~\cite{wang2017chestx} & 0.01 & 0.03 & 0.02 & 0.00 & 0.00 & 0.00 & 0.01 & 0.02 & 0.01 \\
	& Z, Li~\cite{li2018thoracic} & 0.04 & 0.52 & 0.07 & 0.09 & 0.11 & 0.01 & 0.05 & 0.05 & 0.12 \\
	& J, Liu~\cite{liu2019align} & 0.18 & 0.70 & 0.28 & 0.41 & 0.27 & 0.04 & 0.25 & 0.18 & 0.29 \\
	& Ours & \textbf{0.20} & \textbf{0.86} & \textbf{0.48} & \textbf{0.68} & \textbf{0.32} & \textbf{0.14} & \textbf{0.54} & \textbf{0.30} & \textbf{0.44} \\
\hline
\end{tabular}}
\end{center}
\caption{The comparison results of disease localization among the models using 50\% unannotated images and 80\% annotated images. For each disease, the best results are bolded.}
\end{table*}

\begin{table*}
\small
\begin{center}
\resizebox{0.8\textwidth}{!}{
\begin{tabular}{|l|c c c c c c c c c c|}
\hline
T (IoU) & Models & Atelectasis& Cardiomegaly & Effusion & Infiltration & Mass & Nodule & Pneumonia & Pneumothorax & Mean \\
\hline\hline
 	\multirow{3}{*}{0.1} & Z, Li~\cite{li2018thoracic} & 0.59& 0.81 & 0.72 & 0.84 & 0.68 & 0.28 & 0.22 & 0.37 & 0.57 \\
	& J, Liu~\cite{liu2019align}  & 0.39& \textbf{0.90} & 0.65 & \textbf{0.85} & \textbf{0.69} & \textbf{0.38} & 0.30 & 0.39 & 0.60 \\
	
	& Ours & \textbf{0.66} & 0.88 & \textbf{0.79} & \textbf{0.85} & \textbf{0.69} & 0.28 & \textbf{0.40} & \textbf{0.47} & \textbf{0.63}\\
\hline
    \multirow{3}{*}{0.3}& J, Liu~\cite{liu2019align} & 0.34 & 0.71 & \textbf{0.39} & 0.65 & \textbf{0.48} & \textbf{0.09} & 0.16 & 0.20 & 0.38 \\
	& Baseline & \textbf{0.36} & 0.69 & 0.35 & 0.64 & 0.44 & 0.08 & 0.02 & 0.23 & 0.35\\
	& Ours & 0.31 & \textbf{0.79} & 0.37 & \textbf{0.75} & 0.40 & 0.06 & \textbf{0.24} & \textbf{0.27} & \textbf{0.40}\\

\hline
	\multirow{3}{*}{0.5}& J, Liu~\cite{liu2019align} & \textbf{0.19} & 0.53 & \textbf{0.19} & 0.47 & \textbf{0.33} & 0.03 & \textbf{0.08} & 0.11 & 0.24 \\
	& Baseline & 0.18 & 0.51 & 0.14 & 0.47 & 0.27 & 0.03 & 0.01 & 0.12 & 0.22\\
	& Ours & \textbf{0.19} & \textbf{0.71} & 0.14 & \textbf{0.52} & 0.31 & \textbf{0.08} & 0.05 & \textbf{0.13} & \textbf{0.27}\\
\hline
 	\multirow{3}{*}{0.7}& J, Liu~\cite{liu2019align} & 0.08 & 0.30 & \textbf{0.09} & 0.25 & 0.19 & 0.01 & 0.04 & 0.07 & 0.13 \\
	& Baseline & \textbf{0.11} & 0.34 & 0.06 & 0.32 & \textbf{0.20} & 0.01 & 0.00 & 0.06 & 0.14\\
	& Ours & 0.06 & \textbf{0.64} & 0.08 & \textbf{0.38} & 0.19 & 0.01 & \textbf{0.08} & \textbf{0.09} & \textbf{0.19}\\
\hline
\end{tabular}}
\end{center}
\caption{ The comparison results of disease localization among the models using 100\% unannotated images and no any annotated images. For each disease, the best results are bolded.  }
\end{table*}

\begin{table*}
\small
\begin{center}
\resizebox{0.8\textwidth}{!}{
\begin{tabular}{|l|c c c c c c c c c c|}
\hline
T (IoU) & Models & Atelectasis& Cardiomegaly & Effusion & Infiltration & Mass & Nodule & Pneumonia & Pneumothorax & Mean \\
\hline\hline
 	\multirow{3}{*}{0.3}& J, Liu~\cite{liu2019align} & \textbf{0.55} & 0.73 & 0.55 & 0.76 & 0.48 & \textbf{0.22} & 0.39 & \textbf{0.30} & 0.50 \\
 	& Baseline  & 0.47 & 0.84 & 0.65 & 0.82 & 0.33 & 0.04 & \textbf{0.57} & 0.29 & 0.50 \\
    & Ours & 0.49 & \textbf{0.87} & \textbf{0.66} & \textbf{0.88} & \textbf{0.48} & 0.10 & 0.51 & 0.20 & \textbf{0.52} \\
\hline
	\multirow{3}{*}{0.5}& J, Liu~\cite{liu2019align} & \textbf{0.36} & 0.57 & 0.37 & 0.62 & \textbf{0.34} & \textbf{0.13} & 0.23 & 0.17 & 0.35 \\
	& Baseline & 0.27 & 0.76 & 0.39 & 0.58 & 0.24 & 0.02 & 0.39 & \textbf{0.21} & 0.36\\
    & Ours & 0.26 & \textbf{0.80} & \textbf{0.41} & \textbf{0.67} & 0.15 & 0.06 & \textbf{0.42} & 0.18 & \textbf{0.37} \\
\hline
 	\multirow{3}{*}{0.7}& J, Liu~\cite{liu2019align} & \textbf{0.19} & 0.47 & 0.20 & 0.41 & \textbf{0.22} & \textbf{0.06} & 0.12 & \textbf{0.11} & 0.22 \\
	& Baseline & 0.14 & 0.62 & 0.20 & 0.42 & 0.07 & 0.00 & 0.23 & 0.08 & 0.22 \\
	& Ours & 0.18 & \textbf{0.71} & 0.20 & \textbf{0.50} & 0.20 & 0.02 & \textbf{0.29} & 0.06 & \textbf{0.27}\\

\hline
\end{tabular}}
\end{center}
\caption{ The comparison results of disease localization among the models using 100\% unannotated images and 40\% annotated images. For each disease, the best results are bolded.  }
\end{table*}
\begin{figure*}[t]
\begin{center}
\includegraphics[width=0.8\linewidth]{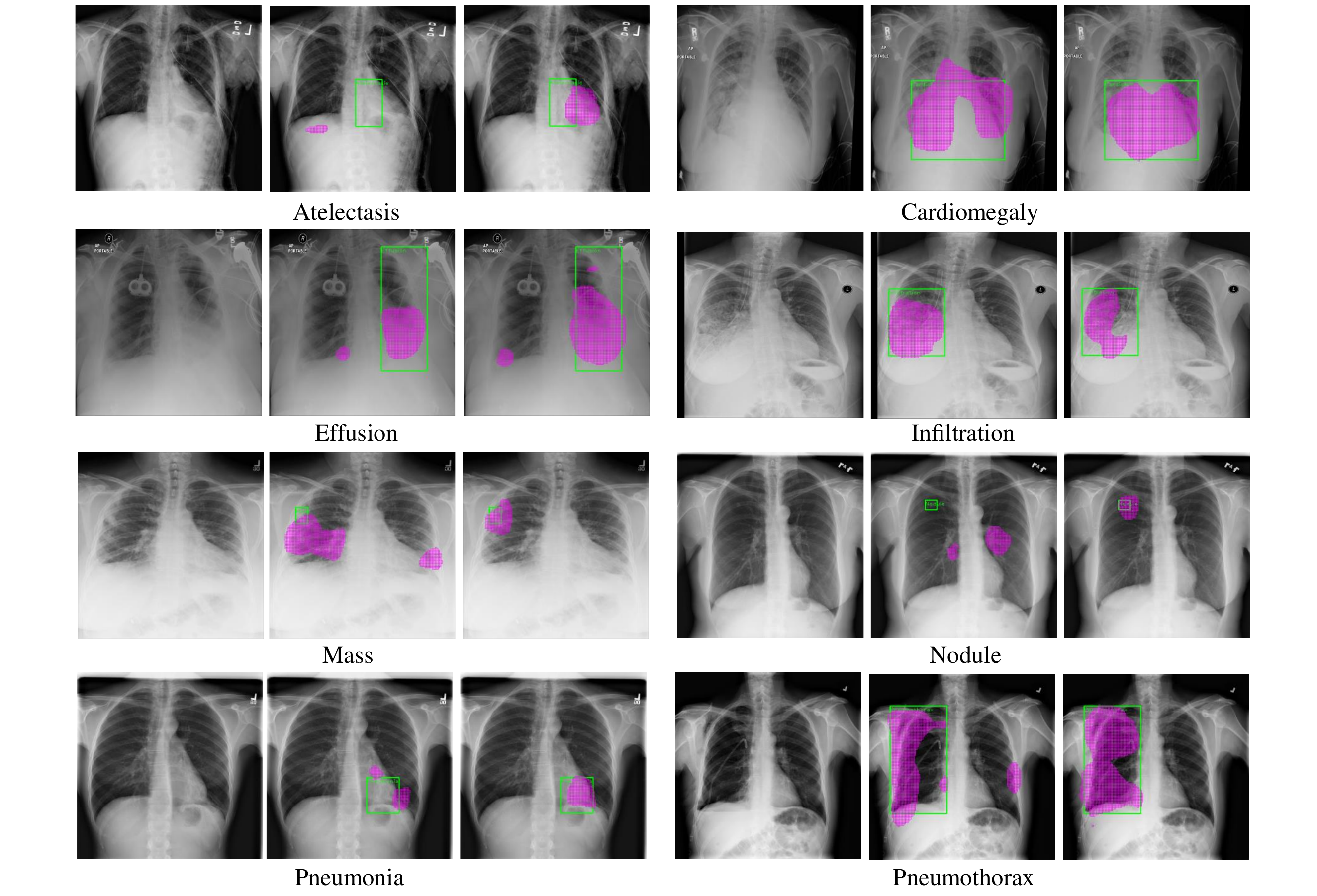}
\caption{Visualization of the predicted results on both the baseline model and our method. The first column shows the original images, the second and third columns show baseline and our method. The green bounding box and red area mean the the ground truth and prediction.}\label{fig:visualization}
\end{center}
\end{figure*}
The graph $G_{k}^{'}$ is calculated on using the graph $G^{'}_{l} \in R^{|p_{i}| \times |p_{j}|}$.
\begin{equation}
\begin{split}
G_{k}^{'} = W^{'}_{l}(G^{'}_{l})
\end{split}
\end{equation}
where $W^{'}_{l}$ is a fully connected layer and
\begin{equation}
\begin{split}
 G_{l}^{'}(l, p) = &D^{'}(P_{l}, P_{p}), \\& l \in \{1,2,...,|p_{i}|\}, p \in \{1,2,...,|p_{j}|\}
\end{split}
\end{equation}
$P_{l}$ is the $l$-th feature patch of $F^{'}_{u}$ and $P_{p}$ is the $p$-th feature patch of $F^{'}_{v}$, respectively.
\subsection{Training Loss}
The overall loss function during the training is a weighted combination of four loss functions,
\begin{equation}\label{eq:loss-all}
\begin{aligned}
L_{all} = w_{1}L_{base} + w_{2}L_{IR} +  w_{3}L_{IK} + w_{4}L_{KR}
\end{aligned}
\end{equation}
where $\sum^4_{i=1}w_{i} = 1$. In our experiments, we always set $w_{i}=0.25, i \in {1,2,..,4}$.

\subsection{Training and Test}
\begin{figure*}[t]
\begin{center}
\includegraphics[width=0.8\linewidth]{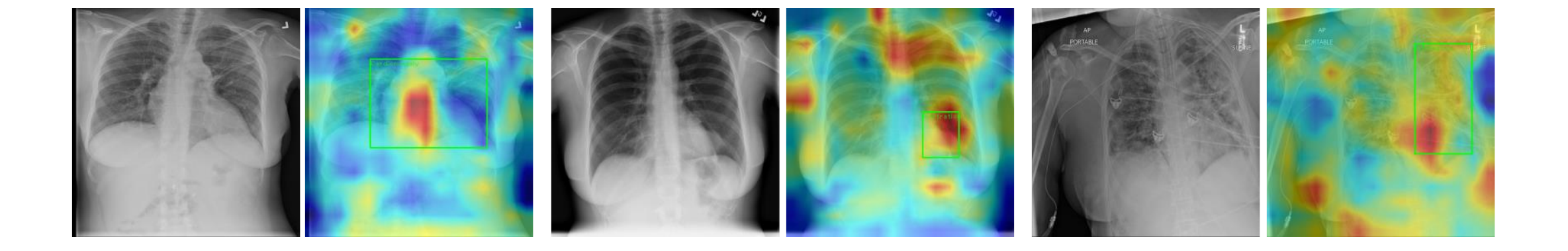}

\caption{Visualization of the generated heatmap and ground truth of our method, where the green bounding box means the ground truth.}\label{fig: visualization}
\end{center}
\end{figure*}
\begin{table*}
\small
\begin{center}
\resizebox{0.8\textwidth}{!}{
\begin{tabular}{|l|c c c c c c c c c c|}
\hline
 Data & Models & Atelectasis& Cardiomegaly & Effusion & Infiltration & Mass & Nodule & Pneumonia & Pneumothorax & Mean \\
\hline\hline
\multirow{9}{*}{0.5\_0.8}& X, Wang~\cite{wang2017chestx} & 0.01 & 0.03 & 0.02 & 0.00 & 0.00 & 0.00 & 0.01 & 0.02 & 0.01 \\
	& Z, Li~\cite{li2018thoracic}  & 0.04 & 0.52 & 0.07 & 0.09 & 0.11 & 0.01 & 0.05 & 0.05 & 0.12 \\
	& J, Liu~\cite{liu2019align} & 0.18 & 0.70 & 0.28 & 0.41 & 0.27 & 0.04 & 0.25 & 0.18 & 0.29 \\
	& Baseline  & \textbf{0.34} & \textbf{1.00} & 0.40 & \textbf{0.68} & 0.11 & \textbf{0.14} & \textbf{0.65} & 0.00 & 0.41 \\
	& IK & 0.22 & 0.82 & 0.36 & 0.56 & 0.32 & \textbf{0.14} & 0.25 & \textbf{0.35} & 0.38 \\
	& IR & 0.24 & 0.82 & 0.40 & 0.56 & 0.32 & 0.07 & 0.38 & 0.30 & 0.39 \\
	& KR & 0.24 & 0.89 & 0.32 & \textbf{0.68} & 0.26 & \textbf{0.14} & 0.21 & 0.30 & 0.38 \\
	& (IR+IK) & 0.20 & 0.86 & \textbf{0.48} & \textbf{0.68} & 0.32 & \textbf{0.14} & 0.54 & 0.30 & \textbf{0.44} \\
	& IR+IK+KR & 0.27 & 0.86 & 0.40 & 0.56 & \textbf{0.37} & \textbf{0.14} & 0.13 & 0.30 & 0.38 \\

\hline
	\multirow{7}{*}{1.0\_0.0}& J, Liu~\cite{liu2019align} & 0.08 & 0.30 & \textbf{0.09} & 0.25 & 0.19 & 0.01 & 0.04 & 0.07 & 0.13 \\
	& Baseline & 0.11 & 0.34 & 0.06 & 0.32 & 0.20 & 0.01 & 0.00 & 0.06 & 0.14\\
	& IK & 0.10 & 0.59 & 0.07 & 0.37 & 0.20 & 0.00 & \textbf{0.13} & 0.06 & \textbf{0.19}\\
 	& GR & 0.06 & 0.61 & 0.07 & 0.28 & 0.14 & 0.00 & 0.05 & 0.08 & 0.16\\
	& IK & 0.09 & 0.63 & 0.06 & 0.36 & \textbf{0.22} & 0.00 & 0.09 & 0.07 & \textbf{0.19}\\
	& IR+IK & 0.06 & \textbf{0.64} & 0.08 & \textbf{0.38} & 0.19 & 0.01 & 0.08 & \textbf{0.09} & \textbf{0.19}\\
	& IR+IK+KR & \textbf{0.12} & 0.51 & 0.07 & 0.36 & \textbf{0.22} & \textbf{0.03} & 0.02 & 0.07 & 0.17 \\
\hline
 	\multirow{7}{*}{1.0\_0.4}& J, Liu~\cite{liu2019align} & \textbf{0.19} & 0.47 & 0.20 & 0.41 & \textbf{0.22} & \textbf{0.06} & 0.12 & 0.11 & 0.22 \\
	& Baseline & 0.14 & 0.62 & 0.20 & 0.42 & 0.07 & 0.00 & 0.23 & 0.08 & 0.22 \\
	& IK & 0.14 & 0.66 & 0.09 & 0.47 & 0.15 & 0.00 & \textbf{0.30} & 0.06 & 0.23\\
	& GR & 0.14 & \textbf{0.75} & \textbf{0.24}  & 0.42 & 0.11 & 0.00  & 0.26 & \textbf{0.12} & 0.25 \\
	& KR  & 0.13 & 0.68 & 0.20 & 0.47 & 0.19 & \textbf{0.06} & 0.17 & 0.08 & 0.25\\
	& IR+IK & 0.13 & 0.72 & 0.13 & 0.43 & 0.20 & 0.00 & 0.23 & 0.06 & 0.24\\
	& IR+IK+KR & 0.18 & 0.71 & 0.20 & \textbf{0.50} & 0.20 & 0.02 & 0.29 & 0.06 & \textbf{0.27}\\
\hline
\end{tabular}}
\end{center}
\caption{ The comparison results of disease localization among the models using three sets of data at T(IoU)=0.7, including 50\% unannotated and 80\% annotated images (0.5\_0.8), 100\% unannotated and no any annotated images (1.0\_0.0), and 100\% unannotated and 40\% unannotated images (1.0\_0.4). For each disease, the best results are bolded.}
\end{table*}
\textbf{Training} All the models are trained on NIH chest X-ray dataset using the SGD algorithm with the Nesterov momentum. With a total of 9 epochs, the learning rate starts from 0.001 and decreases by 10 times after every 4 epochs. Additionally, the weight decay and the momentum is 0.0001 and 0.9, respectively. All the weights are initialized by pre-trained ResNet~\cite{he2016deep} models on ImageNet~\cite{deng2009imagenet}. The mini batch size is set to 2 with the NVIDIA 1080Ti GPU. All models proposed in this paper are implemented based on PyTorch~\cite{paszke2017automatic}.

\textbf{Testing} We also use the threshold of 0.5 to distinguish positive grids from negative grids in the class-wise feature map as described in~\cite{li2018thoracic} and ~\cite{liu2019align}. All test setting is same as~\cite{liu2019align}, we also up-sampled the feature map before two last fully convolutional layers to gain a more accurate localization result.
\section{Experiments}

\subsection{Dataset and Evaluation Metrics}

\textbf{Dataset.} NIH chest X-ray dataset~\cite{wang2017chestx} include 112,120 frontal-view X-ray images of 14 classes of diseases. There are different diseases in each image. Furthermore, the dataset contains 880 images with 984 labeled bounding boxes.
We follow the terms in~\cite{li2018thoracic} and ~\cite{liu2019align} to call 880 images as ``annotated'' and the remaining 111,240 images as ``unannotated''.
Following the setting in ~\cite{liu2019align}, we also resize the original 3-channel images from resolution of $1024\times 1024$ to $512\times 512$ without any data augmentation techniques.

\textbf{Evaluation Metrics.} We follow the metrics used in~\cite{li2018thoracic}. The localization accuracy is calculated by the IoU (Intersection over Union) between predictions and ground truths.
Since it is a coarse-grained task, our localization predictions are discrete small rectangles.
The eight diseases with ground truth boxes is reported in our paper. The localization result is regarded as correct when $IoU > T(IoU)$, where T(*) is the threshold.

\subsection{Comparison with the State-of-the-art}

In order to evaluate the effectiveness of our models for weakly supervised disease detection, we design the experiments on three sets of data and conduct a 5-fold cross-validation. In the first experiment, we use the  50\% unannotated images and 80\% annotated images for training, and test the models with the remaining 20\% annotated images. In the second experiment, we use the 100\% unannotated images and no any annotated images for training, and test the models with all annotated images. In the third experiment, we use the 100\% unannotated images and 40\% annotated images for training, and test the models with remaining 60\% annotated images. Additionally, our experimental results are mainly compared with four methods. The first method is X, Wang~\cite{wang2017chestx}, which proposes a carefully annotated chest X-ray dataset and a unified weakly supervised multi-label image classification and disease localization framework. The second method is Z, Li~\cite{li2018thoracic}, which uses fully convolutional neural network to localize and classify the disease in chest X-ray images. The third method is J, Liu~\cite{liu2019align}, which proposes contrastive learning of paired samples to provide more localization information for disease detection. The last method is our baseline model, which is a unified end-to-end framework that doesn't use our approach to locate and classify the disease.

In the first experiment, we compare the localization results of our model with~\cite{wang2017chestx},~\cite{li2018thoracic} and~\cite{liu2019align}. We can observe that our model outperforms existing methods in most cases, as shown in Table 1. Particularly, with the increase of T(IoU), our model has greater advantages over the reference models. For example, when T(IoU) is 0.3, the mean accuracy of our model is 0.60, and outperforms ~\cite{wang2017chestx}, ~\cite{li2018thoracic} and~\cite{liu2019align} by 0.38, 0.11 and 0.07 respectively. However, when T(IoU) is 0.7, the mean accuracy of our model is 0.44, and outperforms~\cite{wang2017chestx}, ~\cite{li2018thoracic} and~\cite{liu2019align} by 0.43, 0.32 and 0.15 respectively. Overall, the experimental results shown in Table 1 demonstrate that our method is more accurate for disease localization and classification, which provides a great role for clinical practices.

In the second experiment, we train our model without any annotated images comparing the first experiment. Since ~\cite{li2018thoracic} only provides the results when T(IoU) = 0.1, in order to better show the performance of our model, we add an evaluation method of T(IoU) = 0.1. It can be seen that our model outperforms~\cite{li2018thoracic} and~\cite{liu2019align} in most cases, as shown in Table 2. For example, when T(IoU) is 0.1, the mean accuracy of our models is 0.63, which is 0.06 higher than~\cite{li2018thoracic}, and 0.03 higher than~\cite{liu2019align}. Furthermore, when T(IoU) is 0.7, the mean localization result of our model is 0.19, which is 0.06 higher than~\cite{li2018thoracic} and 0.05 higher than~\cite{liu2019align}. Compared with the baseline model, our approach performs better in most classes except for ``Atelectasis'' and ``Nodule". The trend stays the same that at higher T(IoU), our approach demonstrates more advantages over baseline methods. The added unannotated training samples contribute more than the removed annotated ones in those classes, which implies that our approach can better utilize the unannotated samples. The overall results show that even without annotated data used for training, our approach can achieve decent localization results.

In the third experiment, we use more annotated images comparing the second experiment. We compare the localization results of our model with~\cite{liu2019align} in same data setting. It can be seen that our model outperforms~\cite{liu2019align} in most cases, as shown in Table 3. With T(IoU) = 0.3 and 0.7, our model outperforms~\cite{liu2019align} by 0.02 and 0.05 respectively. Similar improvements are achieved comparing the second experiment. Overall, the experimental results demonstrate that our method can improve the performance of models with limited annotated images.

To better demonstrate the final effect of our approach on disease localization and classification, we visualize some of typical predictions of both the baseline model and our method, as shown in Figure 3. The first column shows the original images, the second and third columns show baseline model and our method. The green bounding box and red area mean the ground truth and prediction. It can be seen that our models can predict more accurate in most cases comparing the baseline model. For example, the class ``Atelectasis'' and ``Nodule", the localization reslut of the baseline model is completely inconsistent with the ground truth, but the localization reslut of our method is consistent with the ground truth. It shows that using the structural information of intra-image and inter-image can improve the performance of automatic lesion detection. Additionally, we also visualize the generated heatmap and ground truth of our model, as shown in Figure 4. It can be seen that the proposed method can effectively locate and classify medical images.

\subsection{Ablation Studies}
In this section, we explore the influence of different modules on our method for ablation studies. To evaluate our method more comprehensively, we build 6 models, including the model of the end to end framework (Baseline), the model with the intra-image  knowledge learning (IK), the model with the inter-image relation module (IR), the model with the knowledge reasoning (KR), the model combining the inter-image relation module and the intra-image knowledge learning (IR+IK), the model combining the inter-image relation module, the intra-image knowledge learning and the knowledge reasoning module (IR+IK+KR).

Table 4 shows the results of the three experiments mentioned in section 4.2 at T(IOU)=0.7. It can be seen that our method performs better in most classes except for ``Atelectasis'', ``Effusion" and ``Mass" comparing~\cite{wang2017chestx},~\cite{li2018thoracic} and~\cite{liu2019align}. Furthermore, comparing the baseline model, it can be observed that the performance of our other models are improved in most cases, which shows that our method is effective for improving model performance. However, a model does not always maintain the advantage in the three experiments, for example, the model (IR+IK) achieves the best performance in the data (0.5\_0.8), the model (IK), the model (KR) and the model (IR+IK) achieve the best performance in the data (1.0\_0.0), and the model (IR+IK+KR) achieves the best performance in the data (1.0\_0.4). Overall, the experimental results demonstrate that using structural relational information can improve the performance of models. For different experimental data, our models can achieve different results. It is difficult for us to determine which model is the best, but we can be sure that our method is effective, because no matter what kind of data we use, our models achieve great improvement. Particularly, the method can achieve good localization results even without any annotation images for training.

\section{Conclusion}
By imitating doctor's training and decision-making process, we propose the Cross-chest Graph (CCG) to improve the performance of automatic lesion detection under limited supervision. CCG models the intra-image relationship between different anatomical areas by leveraging the structural information to simulate the doctor's habit of observing different areas. Meanwhile, the relationship between any pair of images is modeled by a knowledge-reasoning module to simulate the doctor's habit of comparing multiple images. We integrate intra-image and inter-image information into a unified end-to-end framework. Experimental results on the NIH Chest-14 dataset demonstrate that the proposed method achieves state-of-the-art performance in diverse situations.

{\small
\bibliographystyle{ieee_fullname}
\bibliography{egbib}
}

\end{document}